\documentclass[letterpaper,10pt,conference]{ieeeconf}
\usepackage{graphicx}
 \usepackage{booktabs}
\usepackage[ruled,vlined]{algorithm2e} 
 \usepackage{pgfplots}
\usepackage{subcaption}   
 \usepackage{pgfplots}
 \usepackage{makecell} 

 \usepackage{bbm}
\usepackage{amssymb}
 \pgfplotsset{compat=1.18}
  \usepackage{amsmath, mathtools}
\usepackage{xcolor}
 \usepackage{multirow}
 \usepackage{svg}
\IEEEoverridecommandlockouts
\title{Designing Latent Safety Filters using Pre-Trained Vision Models}

\author{Ihab Tabbara$^{1*}$, Yuxuan Yang$^{1*}$, Ahmad Hamzeh$^{1}$, Maxwell Astafyev$^{1}$, and Hussein Sibai$^{1}$%
\thanks{$^{1}$Department of Computer Science and Engineering, Washington University in St. Louis, MO 63130, USA.
        {\tt\small \{i.k.tabbara, y.yuxuan, a.h.hamzeh, astafyev, sibai\}@wustl.edu}}%
\thanks{$^{*}$Equal contribution. Ihab Tabbara and Yuxuan Yang are Ph.D. students. Ahmad Hamzeh and Maxwell Astafyev are undergraduate students. Hussein Sibai is an assistant professor.}%
}

\begin{document}

\maketitle

\begin{abstract}
Ensuring safety of vision-based control systems remains a major challenge hindering their deployment  in critical settings. 
Safety filters have gained increased interest as effective tools for ensuring the safety of classical control systems, but their applications in vision-based control settings have so far been limited. Pre-trained vision models (PVRs) have been shown to be effective perception backbones for control in various robotics domains. In this paper, we are interested in examining their effectiveness when used for designing vision-based safety filters. We use them as backbones for classifiers defining failure sets, for Hamilton–Jacobi (HJ) reachability-based safety filters, and for latent world models.
We discuss the trade-offs between training from scratch, fine-tuning, and freezing the PVRs when training the models they are backbones for. We also evaluate whether one of the PVRs is superior across all tasks, evaluate whether learned world models or Q-functions are better for switching decisions to safe policies, and discuss practical considerations for deploying these PVRs on resource-constrained devices.

\end{abstract}

\section{Introduction}

Computer vision plays a critical role in robotics applications,  such as autonomous driving~\cite{waymo2020blog}, manipulation~\cite{cv_for_manip}, and navigation~\cite{CV_for_nav}. It is often the case that the robots do not have direct access to the underlying state of the environment. Instead, they rely on high-dimensional sensory inputs to perceive the world. Images are particularly valuable in this context: they provide rich information about both the robot and the environment states, while being inexpensive and widely available compared to other sensing modalities. However, the safety of vision-based control policies remains a major concern that limits their broader deployment in critical domains. 
Several approaches have 
been proposed to address this challenge
including formal verification~\cite{vision_based_verification_chuchu_stanley_2024,Verification_of_image_based_NN_using_generative_models_JAIS_2022,vision_controllers_verification_Chiao_Sayan_2022}, online monitoring~\cite{monitor_ensembles_Hazem_Torfah_RV_2023,monitoring_using_LLMs_Marco_Pavone_2023}, safe reinforcement learning~\cite{Safe_RL_survey_2015}, and safety filtering~\cite{vision_based_CBF_sarah_dean_CoRL_2021,vcbf,NeRF_CBF_tong_chuchu_icra_2023}.

Safety filters, such as those based on control barrier functions (CBFs)~\cite{CBF_survey_2019,CBF_and_Input_to_State_Safety_for_AD_TCST_2023}, formally guarantee safety of control systems. CBFs can be used to specify controls that guarantee the forward invariance of a set of states. One can then formulate a quadratic program to generate controls that minimally deviate from reference ones while maintaining the invariance of a set that is deemed safe~\cite{cbf}. Moreover, Hamilton–Jacobi (HJ) reachability analysis~\cite{HJ_Bansal_somil_claire_2017}, which we employ in our study, can also be used for safety filtering. Given a user-defined failure set (e.g., the set of points defining an obstacle in the state space), HJ reachability analysis computes the backward reachable set (BRS), i.e., the set of states from which the agent cannot avoid eventually entering the failure set, as the sublevel set of a value function. The complement of the BRS is the maximal controlled forward invariant set~\cite{worst_case_analysis_of_nonlinear_systems_Fialho_TAC_1999}. The analysis also produces a corresponding controller that ensures the forward  invariance of that set by maximizing the value function and consequently steering the system away from the BRS and ensuring safety. However, traditional CBF design and HJ reachability analysis require the system dynamics and formally defined failure set and do not scale beyond few dimensions. 
In complex and dynamic environments, defining a proper state space, obtaining a dynamics model, and formally defining a failure set that has to be avoided becomes challenging. 
Consequently, deep learning approaches have been proposed to address this challenge~\cite{end_to_end_AD_survey_2024,vision_based_CBF_sarah_dean_CoRL_2021,vcbf,NeRF_CBF_tong_chuchu_icra_2023}.

Deep learning approaches for training neural driving policies are either end-to-end or modular. The former 
train models that directly map sensor observations to control inputs, limiting the information loss 
attained in  
modular approaches~\cite{end_to_end_AD_survey_2024}. For vision-based safety filters, most existing methods follow a modular approach, 
assuming the existence of a perception model that maps images to interpretable low-dimensional states over which dynamics are known or are  approximated~\cite{NeRF_CBF_tong_chuchu_icra_2023,vision_based_CBF_sarah_dean_CoRL_2021}. 
Recently, PVRs, which are trained 
 on large-scale datasets \cite{mae}, have been shown to improve the performance and sample complexity of learning end-to-end control policies for various control  tasks~\cite{unsurprising,vc1,r3m}. Such models learn to process images into low-dimensional, non-interpretable, vectors while preserving semantic features. A policy can then be trained on task-specific data to map these features to low-level control, without having to learn common image processing skills from the scarce and costly robotic task-specific data.

 In this work, we 
 systematically evaluate five state-of-the-art PVRs across diverse safety-related tasks and environments. We compare frozen, fine-tuned, and trained-from-scratch variants as backbones for both failure classifiers and HJ reachability-based safety filters, and analyze their performance under different tasks. 
Our findings are as follows: (1) a task-specific representation model can match or even surpass PVRs on safety-control tasks when the PVRs are not fine-tuned; (2) fine-tuning the PVRs during safety-filter training substantially improves collision-avoidance performance; (3) safety filters equipped with DINOv2~\cite{dinov2_pvr} consistently perform well across all tasks; (4) when a PVR is well suited for learning a world model, assessing future-state safety with the world model likely outperforms using a Q function, whereas otherwise a Q function is preferable; and (5) PVR-based safety filters are suitable for real-world deployment.
 To the best of our knowledge, no prior work has evaluated PVRs in the context of safety-critical tasks, nor have existing PVRs been explicitly trained to serve as perception backbones for safety filters rather than reward-driven control policies.

\section{Preliminaries}


\subsection{Hamilton–Jacobi reachability analysis} 
Hamilton–Jacobi (HJ) reachability is a control-theoretic approach  for safety verification and control synthesis~\cite{HJ_Bansal_somil_claire_2017}. 
Consider a dynamical system of the form $s_{t+1} = f(s_t, a_t)$, where $s_t \in S$ and $a_t \in A$. We call the trajectory of the system starting from state $s$ and following a policy $\pi: S \rightarrow A$ by $\xi_s^\pi: \mathbb{R}^{\geq 0} \rightarrow S$.  Given a set of states $\mathcal{F} = \{s \mid h(s) < 0\}$, where $h : S \to \mathbb{R}$, that the user consider a failure set and  wants the system to avoid,  
 HJ reachability analysis  computes the optimal 
 value function $V : S \to \mathbb{R}$, where $V(s) := \sup_{\pi(\cdot)}\inf_{t\geq 0} h(\xi_s^\pi(t))$, which satisfies the fixed-point Bellman equation: $V(s) := \min \left\{ h(s), \max_{a \in A} V(f(s, a)) \right\}$.
 The associated optimal policy is $\pi(s) := \arg \max_{a \in A} V(f(s, a))$. The set of states in the sub-level set of $V$, i.e., $\{s\ |\ V(s) < 0\}$, is called the {\em backward reachable set} (BRS) of the failure set  $\mathcal{F}$. It consists of the states starting from which the system will inevitably reach the failure set and are thus {\em unsafe}. 


For practical implementation in high-dimensional state spaces, we use reinforcement learning (RL) to approximate a time-discounted version of 
the HJ Q-function 
\cite{bridging_HJ_and_RL}.
We employ actor-critic methods such as DDPG \cite{ddpg} and DDQN \cite{ddqn}. 
We optimize the Q-function parameters $\theta$ by minimizing the loss function:
\begin{equation}
L(\theta) = {E}_{(s_t,a_t,s_{t+1})\sim D}\left[(Q_\theta(s_t, a_t) - y_t)^2\right]
\label{eq:safety_q_loss}
\end{equation}
where the target is computed as:
$
y_t = (1-\gamma)h(s_t) + \gamma \min\left\{h(s_t), \max_{a \in A} Q_{\theta}(s_{t+1}, a)\right\}. 
$
When the Q-function is learned, the optimal safety-preserving policy can be retrieved as follows:
$
\pi^*(s) := \arg \max_{a \in A} Q^*(s, a)
\label{eq:optimal_safety_policy}
$
and the corresponding HJ value function evaluated at state $s$ would be:
$
V^*(s) = \max_{a \in A} Q^*(s, a)
\label{eq:value_from_q}
$

\subsection{Pre-trained vision models}
\label{sec:vision_backbone}
We evaluate several popular PVRs which are trained using different  datasets, objectives, and methods: 
VC-1 \cite{vc1}, a Vision Transformer (ViT), was trained on a union of robotic and natural images 
using masked autoencoding (MAE). 
We use its last-layer CLS token as its encoding of the input image.  R3M \cite{r3m}, based on ResNet-50, was trained on the large-scale Ego4D egocentric video dataset with time-contrastive learning, video-language alignment, and an L1 penalty to encourage sparse representations.  DINOv2 \cite{dinov2_pvr}, a self-supervised ViT trained with a teacher–student distillation framework on the LVD-142M dataset (a dataset selected from a corpus comprising benchmarks for image classification, image segmentation, image retrieval, and depth estimation). We consider both the concatenation of last-layer patch embeddings (DINO) and the CLS token output (DINO-CLS) as representations.  Finally, ResNet-50 \cite{resnet} serves as a baseline convolutional encoder trained on ImageNet~\cite{imagenet} and CIFAR-10 with supervised classification objectives.

\subsection{Latent world models and DINO-WM}
World models aim to predict future states of an environment given previous observations and actions, allowing agents to simulate outcomes without direct interaction. 
Latent world models encode high-dimensional observations in 
compact representations 
and model the dynamics 
in the latent space. A latent world model consists of three components: (i) an \textbf{encoder} $z_t \sim \text{enc}_\theta(z_t \mid o_t)$ that maps high-dimensional observations $o_t\in O$ to latent states $z_t\in Z$, (ii) a \textbf{transition model} $z_{t+1} \sim p_\theta(z_{t+1} \mid z_{t-H:t}, a_{t-H:t})$ that predicts future latent states given previous and current latents and actions, and (iii) a \textbf{decoder} $\hat{o}_t \sim q_\theta(o_t \mid z_t)$ that reconstructs observations from the latent states, where $H$ is the temporal history length.

DINO World Model (DINO-WM) \cite{dinowm} builds on this framework by leveraging pretrained visual representations to enable task-agnostic dynamics learning. Unlike approaches such as DreamerV3~\cite{dreamerv3}, which train encoders from scratch with a reward signal to extract task-specific features, DINO-WM employs frozen pretrained encoders, providing rich semantic and spatial priors without reward supervision. Training proceeds as follows: input RGB images are encoded with a frozen pretrained encoder to produce embeddings that define the latent space; the transition model, implemented as a Vision Transformer (ViT), processes sequences of latent embeddings; and action conditioning is achieved by mapping $K$-dimensional action vectors through a multilayer perceptron and concatenating them to the encoded visual features. Proprioceptive data—information about the ego agent’s internal state (e.g., joint positions, velocities)
is incorporated in the same manner. Teacher forcing is used during training, where ground-truth observations are fed instead of predictions, together with a latent consistency loss:
\begin{equation}
\mathcal{L}_{\text{pred}} = \big\|p_\theta(\text{enc}_\theta(o_{t-H:t}), \phi(a_{t-H:t})) - \text{enc}_\theta(o_{t+1}) \big\|_2^2,
\end{equation}
where $\phi$ is the action encoder. This design allows DINO-WM to exploit pretrained visual priors while learning predictive dynamics directly in latent space, providing strong representations without requiring reward-based supervision. In our work, we adapt DINO-WM to use any of R3M, Resnet, VC1, DINO, and DINO-CLS as the pre-trained vision encoders.

\section{Related Work}

\noindent\textbf{PVR as backbones for control} Several works have shown the benefits of using PVRs as backbones for control policies. The authors in \cite{unsurprising} showed that frozen PVR backbones can be competitive with, and sometimes outperform, policies with access to ground truth states in various benchmarks. The authors in \cite{vc1} further compared frozen and fine-tuned PVRs across control tasks, finding no single dominant model and introducing VC1, which outperforms others on average. In this work, we evaluate state-of-the-art PVRs as backbones for safety filters, comparing frozen, fine-tuned, and models trained from scratch across multiple tasks.

\noindent\textbf{Vision-based safety filters} 
HJ-based safety filters in latent space have recently been explored. The works of~\cite{andrea_latentsafety_uncertainty} and~\cite{andrea_latentsafety} demonstrate this using DINO-WM~\cite{dinowm} with a frozen DINO backbone, as well as DreamerV3~\cite{dreamerv3}, which instead learns a task-specific image encoder during training. Aside from HJ-based latent safety filters, CBFs have also been used in vision-based settings ~\cite{BarrierNet_DanielaRus_2023, NeRF_CBF_tong_chuchu_icra_2023}. 

\section{Methodology}
In this section, we first describe the training of DINO-WM, latent-space failure classifiers, and safety filters. We then outline our evaluation methodology for the latter two.


\subsection{Learning latent dynamics}
We adapt DINO-WM \cite{dinowm} to learn dynamics as well as proprioception and action encoders across five PVRs: VC1, R3M, ResNet, DINO, and DINO-CLS. For each environment, we also train a new vision encoder, denoted WM-R, by randomly initializing a ResNet architecture and training DINO-WM end-to-end, thereby learning the representation and the dynamics jointly. We set the history length to $H=3$ and train each model for 100 epochs on datasets we collected with the environment’s reference controller.

\subsection{Learning the failure classifier in latent space}
 For the open-loop evaluation, we learn a function $h$ defined over the latent space, 
whose $0$-sublevel set specifies the failure region, i.e., $\mathcal{F}_{o_f}=\{{o_f} \in O \mid h(\text{enc}_\theta({o_f})) \leq 0\} $. We train $h$ as a 
multi-layer perceptron (MLP), and employ a hinge-style loss 
to encourage margin separation between the failure region and the non-failure region:
$
    \mathcal{L}_{h} =
    \sum_{o_f\in\mathcal{F}_{o_f}}\sigma(\alpha-h(\text{enc}_\theta(o_f)))
     + \sum_{\bar{o}_f\in\bar{\mathcal{F}}_{o_f}}\sigma(\alpha+h(\text{enc}_\theta(\bar{o}_f))),
$
%
%
where $\sigma$ is the ReLU function and the hyperparameter $\alpha > 0$ to prevent the neural network from predicting single value, $o_f$ corresponds to the observation inside the failure set $\mathcal{F}_{o_f}$, and $\bar{o}_f$ corresponds to the observation inside the compliment set of the failure set $\bar{\mathcal{F}}_{o_f}$. We access the failure labels of the observations using the simulator of each environment.

To improve the learned boundary, we incorporate a
\emph{gradient penalty} term on interpolated latent features between failure and non-failure states:
$
    \mathcal{L}_{\text{gp}} =
    \mathbb{E}_{\tilde{z}} \big[
    (\|\nabla_{\tilde{z}} h(\tilde{z})\|_2 - \lambda)^2
    \big],
$

%
\noindent where $\tilde{z} = \alpha h(\text{enc}_\theta({o_f})) + (1-\alpha) h(\text{enc}_\theta(\bar{o}_f))$, with
$\alpha \sim \mathcal{U}(0,1)$ and $\lambda$ being the target gradient norm.
The final training objective is $\mathcal{L} = \mathcal{L}_{h} + \beta \mathcal{L}_{\text{gp}}$,
where $\beta$ controls the strength of the
penalty. This setup encourages $h$ to
be both discriminative and smooth near failure set boundary.
\label{sec:learning_h}

\subsection{Learning the HJ value function as the safety filter for closed-loop evaluation}
\label{sec:learning_hj}
We train the HJ value function as the safety filter using DDPG for environments with continuous action spaces and DDQN for environments with discrete action spaces and optimize the loss in (\ref{eq:safety_q_loss}). When the simulator provides a ground-truth distance function, we use it directly; otherwise, if it only provides boolean failure labels, we rely on the output of the learned classifier $h(\text{enc}_\theta(o))$ introduced in Section~\ref{sec:learning_h}.

For closed-loop evaluation, we use a switching scheme that can operate in either \texttt{critic-only} or \texttt{dynamics lookahead} mode presented in Algorithm \ref{alg:switching_algos}. Both variants decide whether to follow the nominal policy $\pi_{\rm nom}$ or switch to the safe policy $\pi_{\rm safe}$ generated by the HJ reachability analysis. In  \texttt{critic-only} mode, the decision relies directly on the safety critic $Q$. In  \texttt{dynamics lookahead} mode, the world model $\hat{f}_\eta$ predicts the next latent state under the nominal action, and the critic evaluates its safety based on this prediction.

\vspace{-0.25cm}
\begin{algorithm}[h]
\caption{Switching between reference and safe controller}
\hspace{-0.3cm}\KwIn{latent state $z_t$, nominal policy $\pi_{\rm nom}$, safety critic $Q$, world model $\hat{f}_\eta$, margin $\tau \geq 0$, method $\in \{\texttt{Critic-only}, \texttt{dynamics lookahead}\}$}
\hspace{-0.3cm}\KwOut{execute action $a_t$}

$a_{\rm nom} \gets \pi_{\rm nom}(z_t)$\;

\eIf{method = \texttt{dynamics lookahead}}{
    $\hat{z}_{t+1} \gets \hat{f}_\eta(z_t, a_{\rm nom})$; $p \gets \max_{a} Q(\hat{z}_{t+1}, a)$\;
}{
    $p \gets Q(z_t, a_{\rm nom})$\;
}

\eIf{$p < \tau$}{
    $a_t \gets \pi_{\rm safe}(z_t) := \arg\max_{a} Q(z_t, a)$\;
}{
    $a_t \gets a_{\rm nom}$\;
}

\label{alg:switching_algos}
\end{algorithm}
\vspace{-0.65cm}

\subsection{Fine-tuning PVRs and learning representation models}

When learning the failure classifier in latent space, we first train while freezing the PVR backbones and then repeat the experiment with fine-tuning them. In the latter case, the loss backpropagates through all the encoders from DINO-WM. We follow the same steps when training the HJ value function: one variant with frozen backbones and another with fine-tuned backbones. Importantly, when fine-tuning the backbones for HJ training, the learned dynamics can no longer be used since they were trained on the frozen representation, which changed. In addition, for both the failure classifier and HJ value functions, we train ViTs from scratch to test whether the task-specific losses alone are sufficient to learn a good representation. We call the learned vision representation model in these tasks ``Scratch". Altogether, for each of the failure classifier in latent space and HJ tasks we obtain 13 model variants: five frozen PVR backbones, five fine-tuned PVR backbones, one frozen WM-R backbone (a randomly initialized model with ResNet architecture trained jointly with dynamics when training DINO-WM), one unfrozen WM-R model, and one  ``Scratch" model, referring to a randomly initialized ViT model that is trained end-to-end for each task, where the loss is backpropagated through the backbone to jointly learn both the representation and either the latent failure classifier or the HJ value function.

\subsection{Evaluation metrics}

\subsubsection{Classification with learned failure classifier}
We probe each backbone’s ability to distinguish failure from non-failure states with the learned failure classifiers. First, we compute the correlation between the learned $h$ and the ground-truth distance function provided by the simulator. When the ground truth distance function is not accessible, we compute the correlation with the binary failure labels.  A high correlation indicates that the learned
representation is sufficiently rich to capture how $h$ evolves along a
trajectory (i.e., whether the agent is moving closer to an obstacle at any given timestep). We also probe the PVRs by showing classification accuracy for the failure and non-failure states.
\subsubsection{HJ in Closed Loop}
We evaluate two conditions: (1) \textbf{No Safety Filter}, where the nominal policy acts alone, and (2) \textbf{HJ Safety Filter Enabled}, where the reference policy is accompanied by the HJ safety filter. Each evaluation consists of 50 independent episodes with random initial states, which are kept fixed across all backbones and switching methods for fair comparisons. We report the \textbf{success rate} (percentage of episodes that reach the task goal),
\textbf{violation count} (average number of states where a failure state is reached, e.g., collision), inference time, and the size of each safety filter.

\subsubsection{Saliency maps}
To interpret which regions of the visual input most influence the
safety filter, we generated occlusion-based saliency maps.
Given an input observation,
we replace square patches of the image with black or white patches. For each modified image, we measure the change in predicted HJ value relative to the original
prediction. By sliding an occlusion window across the image with stride $k$, replacing each patch, and measuring the resulting change in the HJ value, we generate a saliency map.



\section{Experimental Setup}
We evaluate the above approach in four simulated environments. In each environment, we describe specific tasks, failure conditions, proprioception of the agent, and data collected to train the DINO-WM and the failure classifiers. All models were trained using either an NVIDIA A40 or RTX~5090, 200\,GB of system RAM and between 2-16 CPUs depending on the environment. Each world model was trained on a A40, requiring 1–2 days per environment–backbone pair on average. HJ training was conducted on an RTX~5090, taking 1–2 days on average per backbone.


\subsection{Environments}
\noindent\textbf{ManiSkill (UnitreeG1PlaceAppleInBowl):} We simulate a humanoid (UnitreeG1) robot in ManiSkill \cite{maniskill2} to pick up an apple, with a random initial position, and place it in a bowl. The observation is an RGB image captured by a camera mounted on the robot's head. The proprioception $\mathbf{x}$ contains the positions and velocities of the 25 joints of the robot. $o_{f} \in \mathcal{F}_{o_f}$ if the robot's hand comes into contact with the bowl. We trained a PPO policy on the full state which includes $\mathbf{x}$ and environment information, such as the bowl's and the apple's position. In this task, where only binary collision labels are available from the simulator, we first learn the failure classifier as described in Section~\ref{sec:learning_h}, then use it to learn the HJ safety filter as described in Section~\ref{sec:learning_hj}. We collect and label a dataset of 3,000 episodes using the trained PPO policy.

\noindent\textbf{Dubins Car (2D Navigation):} We simulate a Dubins car moving in a 2D plane at constant speed with two obstacles. The observation is an overhead RGB image of the environment. The task is to navigate to a goal location without colliding with obstacles. The proprioception is $\mathbf{x} = (x, y, \theta)$. We exclude the proprioception from all training, as $(x, y, \theta)$ can be inferred from the RGB image. The reference is a PID controller tracking the distance to the obstacle. We determine if $o_{f} \in \mathcal{F}_{o_f}$ using the ground-truth distance to the closest obstacle from the simulator. We collect and label a dataset of 2,000 episodes using the PID controller. 

\noindent\textbf{Safety Gymnasium CarGoal:} 
An agent must navigate to a target goal position in an arena that contains hazardous areas in Safety Gymnasium~\cite{safety_gymnasium}. The visual observation is an egocentric view from the car, and the proprioception $\mathbf{x} \in \mathbb{R}^{24}$ contains information about the agent, such as angular velocities, gyro, etc. We train a nominal policy to reach goals using DreamerV3 \cite{dreamerv3} without consideration of any obstacles (no cost signal is given while learning the nominal policy). We define $o_{f} \in \mathcal{F}_{o_f}$ if a collision occurs between the car and any hazard, which is determined using LIDAR measurements from the environment. We collect and label a dataset of 2,000 episodes using the policy obtained from DreamerV3.

\noindent\textbf{CARLA:} 
In CARLA \cite{CARLA_original_paper_CoRL_2017}, we used a scenario in which the ego vehicle followed a leading non-ego. The lead vehicle implemented a lane-following PID-driven controller, while the ego car uses a noisy PID controller (to generate trajectories with collisions). For simulated trajectories, vehicle models and spawn points are randomly chosen from CARLA's predefined vehicle blueprints and Town10 environment, respectively. Raw observations are RGB images captured from a camera mounted at the center of the windshield. 
The proprioception is $\mathbf{x} =(x,y,z,v,\theta,\psi,\phi)$, which are respectively the ego vehicle's position (x, y, z), velocity, and orientation (pitch, yaw, roll). $o_{f} \in \mathcal{F}_{o_f}$ when two vehicles collide. We collect and label a dataset of 2,000 episodes using the noisy PID controller.

\section{Results}
In this section, we present and analyze our results by addressing a set of research questions.

\renewcommand\theadfont{\normalfont\bfseries}

\begin{table}[ht]
\definecolor{darkgreen}{RGB}{0,150,0}
\centering
\setlength{\tabcolsep}{6pt} 
\renewcommand{\arraystretch}{1.1} 
\tiny 
\begin{tabular}{l|cc cc cc}
\toprule
\multirow{2}{1.25cm}[-1.5ex]{\centering \textbf{PVR}} & \multicolumn{2}{c}{\textbf{ManiSkill}} & \multicolumn{2}{c}{\textbf{Dubins}} \\
\cmidrule(lr){2-3} \cmidrule(lr){4-5} \cmidrule(lr){6-7}
& \textbf{FT} & \textbf{No FT} & \textbf{FT} & \textbf{No FT} \\
\midrule
DINO-CLS & 0.449 & \underline{0.451} & 0.908 & \underline{0.976} \\
DINO & 0.617 &\textbf{\textcolor{darkgreen}{\underline{0.770}}} & 0.949 & \textbf{\textcolor{darkgreen}{\underline{0.981}}} \\
VC1 & \underline{0.489} & 0.461 & 0.956 & \underline{0.958}\\
R3M& \underline{0.542} & 0.514 & 0.942 & \underline{0.963}  \\
ResNet & \textbf{\textcolor{darkgreen}{\underline{0.620}}} & 0.451 & 0.966 & \underline{0.974} \\
WM-R & \underline{0.517} & 0.442 & \textbf{\textcolor{darkgreen}{\underline{0.978}}} & 0.935 \\
Scratch & 0.494 & - & 0.922 & - \\
\midrule
\multirow{2}{*}{\textbf{}} & \multicolumn{2}{c}{\textbf{CarGoal}} & \multicolumn{2}{c}{\textbf{Carla}} \\
\cmidrule(lr){2-3} \cmidrule(lr){4-5} \cmidrule(lr){6-7}
& \textbf{FT} & \textbf{No FT} & \textbf{FT} & \textbf{No FT} \\
\midrule
DINO-CLS & 0.899 & \underline{0.901} & \underline{0.600} & 0.238 \\
DINO & \textbf{\textcolor{darkgreen}{\underline{0.927}}} & 0.923 & \underline{0.314} & 0.305 \\
VC1 & \underline{0.789} & 0.785 & 0.428 & \underline{0.556}  \\
R3M& \underline{0.924} & 0.871 & \underline{0.496} & 0.488  \\
ResNet & \underline{0.902} & 0.899 & \underline{0.551} & 0.395 \\
WM-R & 0.909 & \textbf{\textcolor{darkgreen}{\underline{0.930}}} & \textbf{\textcolor{darkgreen}{0.615}} & \textbf{\textcolor{darkgreen}{\underline{0.638}}} \\
Scratch & 0.801 & - & 0.436 & - \\
\bottomrule
\end{tabular}
\caption{Correlation scores for failure classifiers across environments and backbones. Backbones in \textbf{\textcolor{darkgreen}{green}} are best backbone for each task in the case of fine-tuning (FT) and freezing the backbone (NoFT) while training the latent failure classifer. \underline{Underlined} backbones correspond to whether the FT or the Non-FT version performed better for each task.}\label{tab:corr}
\vspace{-0.4cm}
\end{table}

\begin{table}[ht]
\definecolor{darkgreen}{RGB}{0,150,0}
\centering
\setlength{\tabcolsep}{6pt} 
\renewcommand{\arraystretch}{1.1} 
\tiny
\begin{tabular}{l| c| l|cc cc cc}
\toprule
\multirow{2}{0.8cm}[-1.5ex]{\centering \textbf{Method}} & \multirow{2}{*}{ \textbf{\makecell{Fine-tuned\\PVR}}} & \multirow{2}{0.8cm}[-1.5ex]{\centering \textbf{PVR}} & \multicolumn{2}{c}{\textbf{ManiSkill}} & \multicolumn{2}{c}{\textbf{Dubins}} & \multicolumn{2}{c}{\textbf{CarGoal}}\\
\cmidrule(lr){4-5} \cmidrule(lr){6-7} \cmidrule(lr){8-9}
&& & \textbf{Vio.} & \textbf{Succ.} & \textbf{Vio.} & \textbf{Succ.} & \textbf{Vio.} & \textbf{Succ.} \\
\midrule
\multirow{12}{*}{Critic-only} & \multirow{6}{0.8cm}[-1.5ex]{\centering No} & DINO-CLS &\textbf{\textcolor{darkgreen}{1.28}} & \textbf{\textcolor{darkgreen}{0.34}} & 4.96 & 0.98 & 48.62 & 0.44 \\
& & DINO & 3.52 & 0.10 & 9.80 & 0.98 & 57.70 & 0.20\\
& & VC1 & 4.76 & 0.02 & 8.92 & 0.98 & 53.48 & 0.38\\
& & R3M& 1.88 & 0.38 & 8.44 & 0.98 & 73.42 & 0.86\\
& & ResNet & 5.40 & 0.32 & 8.76 & 0.98 & 54.82 & 0.42\\
& & WM-R & 3.82 & 0.18 & \textbf{\textcolor{darkgreen}{4.66}} & \textbf{\textcolor{darkgreen}{0.98}} & \textbf{\textcolor{darkgreen}{48.26}} & \textbf{\textcolor{darkgreen}{0.32}}\\
\cmidrule(lr){2-9}
& \multirow{6}{0.8cm}[-1.5ex]{\centering Yes} & DINO-CLS & \textbf{\underline{\textcolor{darkgreen}{0.42}}} & \textbf{\underline{\textcolor{darkgreen}{0.44}}} & \underline{\textbf{\textcolor{darkgreen}{0.60}}} & \underline{\textbf{\textcolor{darkgreen}{0.98}}} & \underline{30.68} & \underline{0.26}\\
& & DINO & 4.62 & 0.36 & \underline{2.58} & \underline{0.98} & 35.78 & 0.26\\
& & VC1 & \underline{1.74} & \underline{0.00} & \underline{7.26} & \underline{0.98} & \underline{\textbf{\textcolor{darkgreen}{19.40}}} & \underline{\textbf{\textcolor{darkgreen}{0.24}}}\\
& & R3M& \underline{0.44}& \underline{0.50} & \underline{7.90} & \underline{0.98} & 48.52 & 0.32\\
& & ResNet & 5.30 & 0.32 & \underline{5.50} & \underline{0.98} & \underline{19.74} & \underline{0.20}\\
& & WM-R & 8.10 & 0.42 & \underline{4.50} & \underline{0.98} & \underline{35.70} & \underline{0.06}\\
\midrule
\multirow{6}{*}{\shortstack{Dynamics \\ lookahead}} & \multirow{5}{0.8cm}[-1.5ex]{\centering No} & DINO-CLS & 3.64 & 0.44 & 5.68 & 1.00 & 68.82 & 0.44\\
& & DINO & \underline{\textbf{\textcolor{darkgreen}{1.86}}}& \underline{\textbf{\textcolor{darkgreen}{0.54}}} & 9.44 & 0.98 & \underline{\textbf{\textcolor{darkgreen}{35.16}}} & \underline{\textbf{\textcolor{darkgreen}{0.22}}}\\
& & VC1 & 4.62 & 0.34 & 8.68 & 0.98 & 77.36 & 0.58\\
& & R3M& 8.14 & 0.32 & 8.18 & 0.98 & \underline{35.60} & \underline{0.00}\\
& & ResNet & \underline{3.74} & \underline{0.50} & 8.18 & 0.98 & 60.36 & 0.98\\
& & WM-R & \underline{3.56} & \underline{0.54} & \textbf{\textcolor{darkgreen}{4.72}} & \textbf{\textcolor{darkgreen}{0.98}} & 79.62 & 0.42\\
\midrule
Critic-only&-& Scratch & 2.82 & 0.02 & 5.96 & 0.98 & 36.04 & 0.28\\
&& Nominal & 4.98 & 0.36 & 8.18 & 0.98 & 61.06 & 1.00\\
\bottomrule
\end{tabular}
\caption{Success rates (Succ.) and total violations (Vio.) for safety filters with different backbones across tasks.  "Method" indicates whether the critic-only or the dynamics lookahead algorithm was used in the evaluation. \textbf{\textcolor{darkgreen}{green}} indicates the lowest violation among all backbones per method and fine-tune setting, while \underline{underline} denotes the lowest violation for each PVR across different methods and fine-tune settings. We also highlight the corresponding success rates of the highlighted lowest violation.}\label{tab:closeloop}
\vspace{-0.3cm}
\end{table}

\subsection{Are PVRs sufficient as backbones for safety filtering, or do we train representations from scratch for every task?}
The question can be divided into two parts: Are PVRs sufficient for (1) distinguishing safe from unsafe observations and (2) training safety filters?

\begin{figure}[h]
\centering
\includegraphics[width=0.45\textwidth]{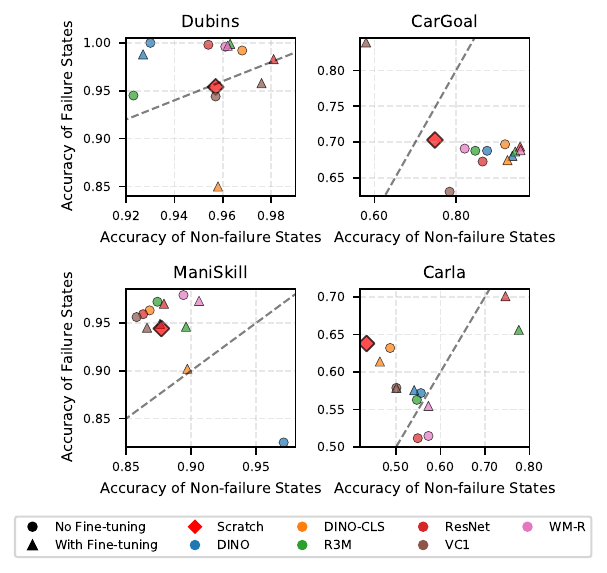}
\caption{Probing classification accuracy of the failure set $\mathcal{F}$ and the non-failure set $\bar{\mathcal{F}}$ for failure classifiers with and without backbone fine-tuning. Dashed lines denote $y=x$.}
\vspace{-0.7cm}
\label{fig:acc}
\end{figure}

We first evaluate which representation model is most suited as the vision backbone for the failure classifiers in latent space. As shown in Table~\ref{tab:corr}, when backbones are not fine-tuned, failure classifiers using WM-R achieve the highest correlation scores on average, and those using Scratch underperform. After fine-tuning the PVRs, they outperform both failure classifiers with WM-R and Scratch backbones. From Figure~\ref{fig:acc}, we observe that failure classifiers with WM-R as the backbone achieve accuracies comparable to those using non-fine-tuned PVRs, while the Scratch backbone performs noticeably worse than both WM-R and the PVRs.

To address the second question, we compare the PVRs with both the Scratch and the WM-R backbones. Since the dynamics model is unavailable for the Scratch backbone, we compare it only against safety filters evaluated using the \texttt{Critic-only} method. In Table~\ref{tab:closeloop}, the results show that, compared to filters without backbone fine-tuning and using the \texttt{Critic-only} method, the safety filters using the Scratch and WM-R backbones perform better on average. In particular, in the Dubins car experiment, the safety filters using the Scratch backbones achieve the lowest number of violations in CarGoal and the third lowest in both ManiSkill and Dubins compared to the safety filters with non-fine-tuned PVR backbones. Also, WM-R, which learns its representation during world-model training, achieves the lowest average number of violations on both Dubins car and CarGoal PVRs and performs relatively well in Maniskill compared to the non-fine-tuned PVRs. These findings suggest that when the backbone is not fine-tuned during safety-filter training, a task-specific representation such as Scratch and WM-R can match or even surpass PVRs trained on large datasets such as the ones we consider. 
Nevertheless, across all tasks, safety filters with fine-tuned PVRs generally reduce violations more than those with the Scratch and WM-R backbones. For example, Table~\ref{tab:closeloop} shows that after fine-tuning the DINO-CLS backbone during the HJ training, its safety filter achieves the safest performance across all tasks compared to safety filters using WM-R and Scratch as backbones. 

\begin{figure}[h]
\vspace{-0.1cm}
    \centering
    \subcaptionbox{DINO (FT)\label{subfig:dino_ft_brs}}{
    \includegraphics[width = .20\linewidth]{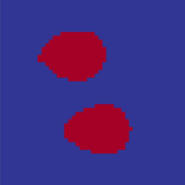}
    }
    \subcaptionbox{DINO\label{subfig:dino_brs}}{
    \includegraphics[width = .20\linewidth]{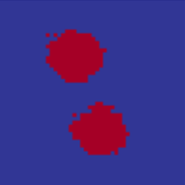}
    }
    \subcaptionbox{Scratch\label{subfig:scratch_brs}}{
    \includegraphics[width = .20\linewidth]{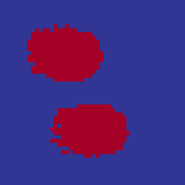}
    }
    \subcaptionbox{GT \label{subfig:gt_brs}}{
    \includegraphics[width = .20\linewidth]{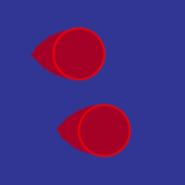}
    }
\caption{BRS recovered from  HJ value functions learned in latent space of different PVRs. $\theta$ is fixed to 0, and \textcolor{red}{red} pixels denote states within the BRS (i.e., $V(z)<0$). GT denotes the ground truth BRS. } 
\vspace{-0.4cm}
\label{fig:brs}
\end{figure}

To understand why Scratch performs better compared to frozen PVRs but subpar compared to fine-tuned ones while training the HJ, we visualize the learned BRS for the Dubins task in Figure~\ref{fig:brs}. It can be seen how, for the HJ value function using DINO as the backbone, the recovered BRS without fine-tuning is worse than that recovered from the filter trained with the Scratch backbone. On the other hand, the recovered BRS from the HJ value function after fine-tuning the DINO backbone is substantially more accurate. Since HJ training time is comparable for frozen and fine-tuned backbones, fine-tuning is better than training new backbones from scratch for each task, as we do when training ``Scratch" and ``WM-R".

\subsection{Should we fine-tune the PVR when learning the safety filters?}


Table~\ref{tab:closeloop} shows that during the closed-loop evaluation, only two of the 18 safety filters exhibit an increase in the average number of violations when their backbones are fine-tuned, which means fine-tuning the backbone during training generally improves safety filters’ collision-avoidance performance. This finding is consistent with prior work on robot control tasks that do not consider safety \cite{vc1}. However, fine-tuning the backbones can render the safety filter overly conservative. On the CarGoal task, all safety filters (except the DINO-based filter) experience a decrease in success rate when their backbones are fine-tuned. Notably, safety filters using WM-R derive a marginal decrease in the number of violations from fine-tuning the backbone (Dubins, CarGoal), or a significant increase (ManiSkill). 

\subsection{Does any PVR consistently outperform the others across all tasks?}
\begin{figure}[h]
    \centering
    \subcaptionbox{DINO-CLS}{
    \includegraphics[width = .2\linewidth]{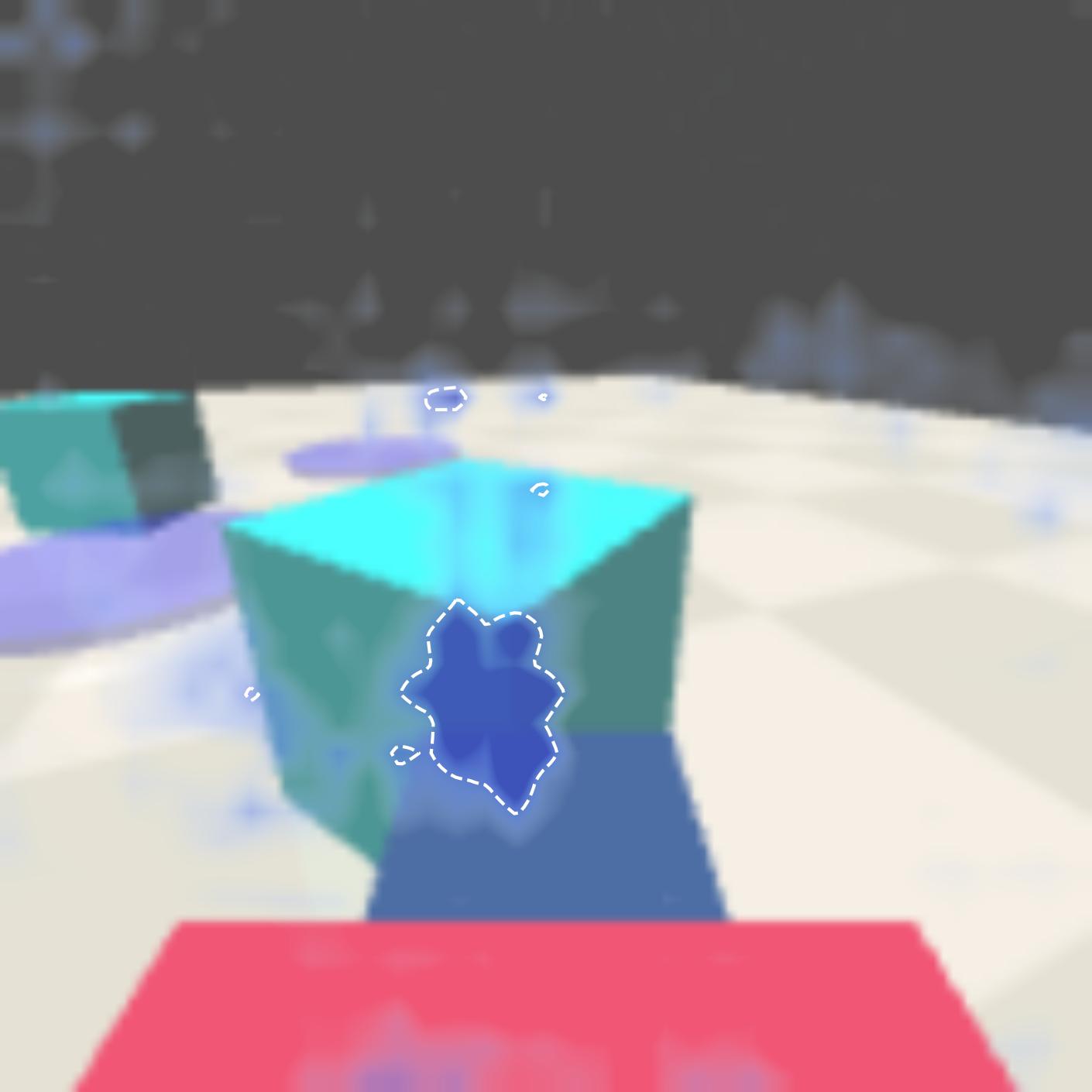}
    }
    \subcaptionbox{DINO}{
    \includegraphics[width = .2\linewidth]{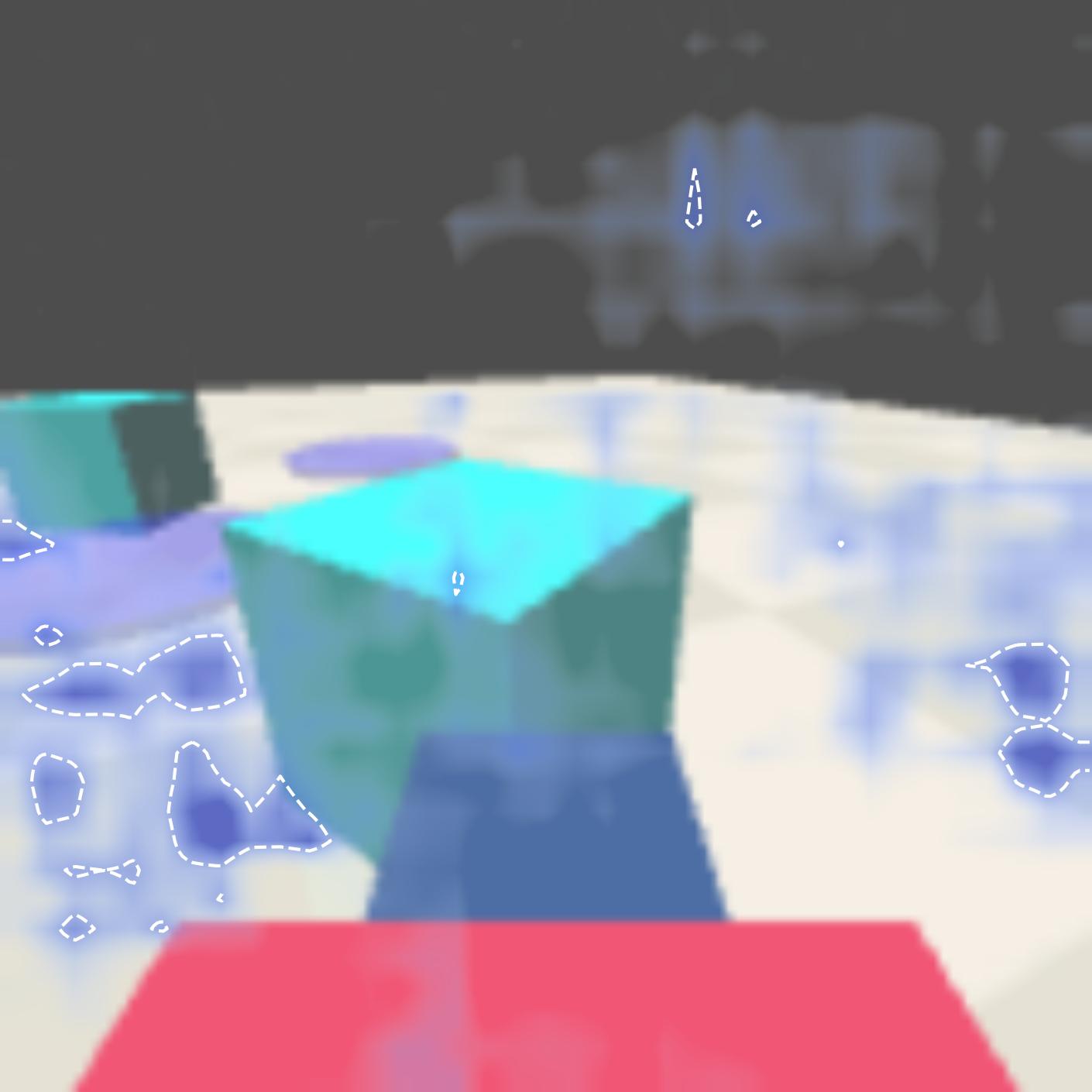}
    }
    \subcaptionbox{VC1}{
    \includegraphics[width = .2\linewidth]{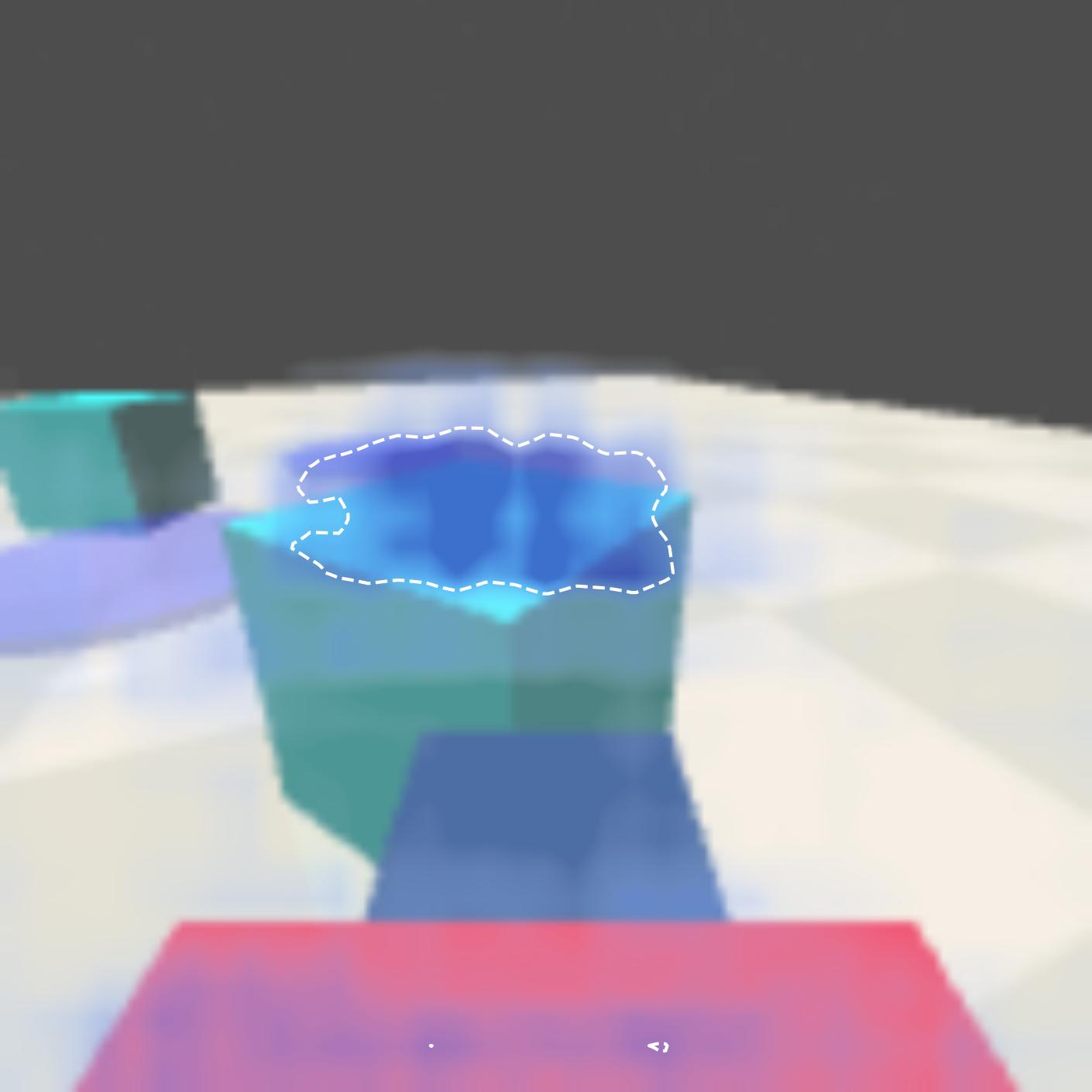}
    }
    \\
    \subcaptionbox{R3M}{
    \includegraphics[width = .2\linewidth]{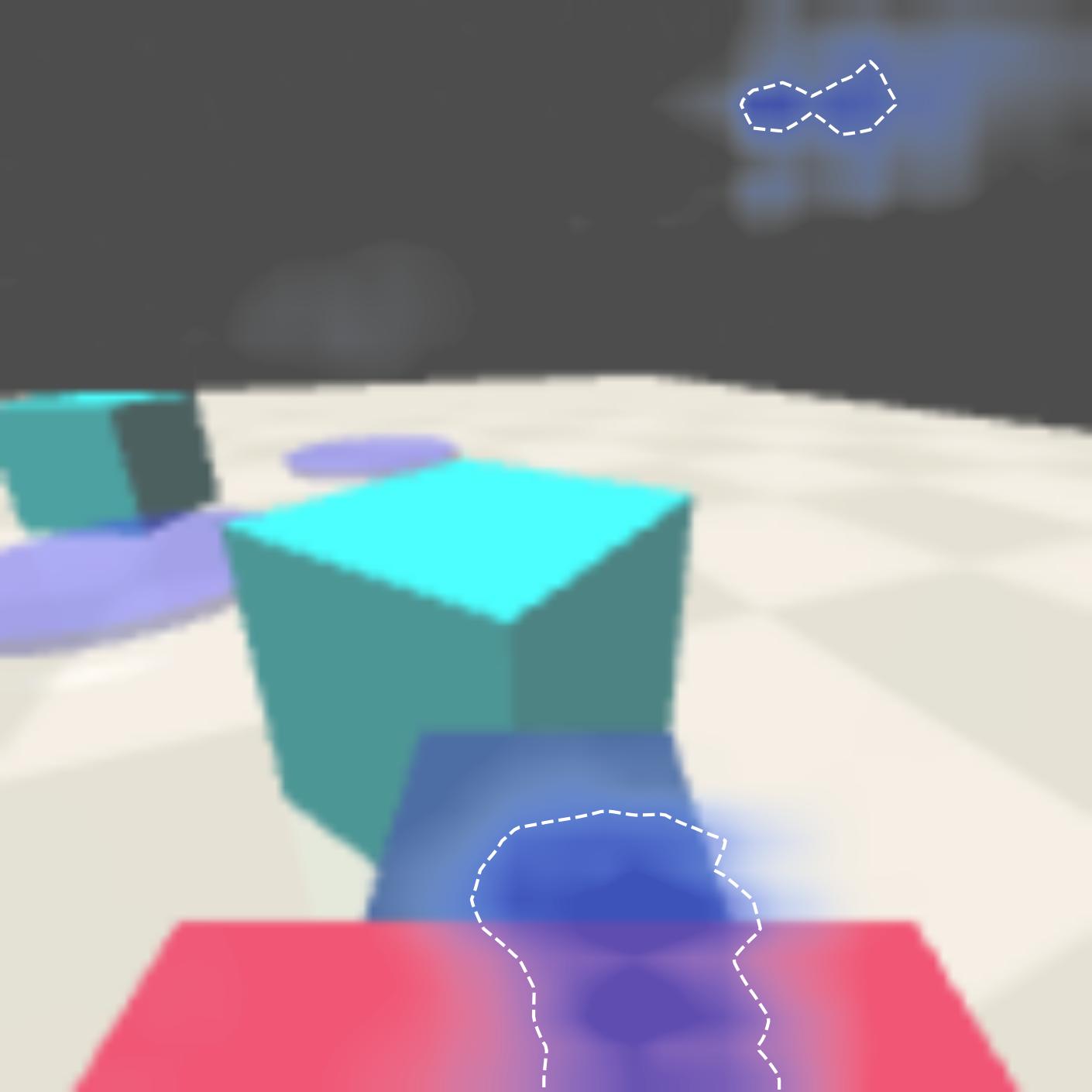}
    }
    \subcaptionbox{ResNet}{
    \includegraphics[width = .2\linewidth]{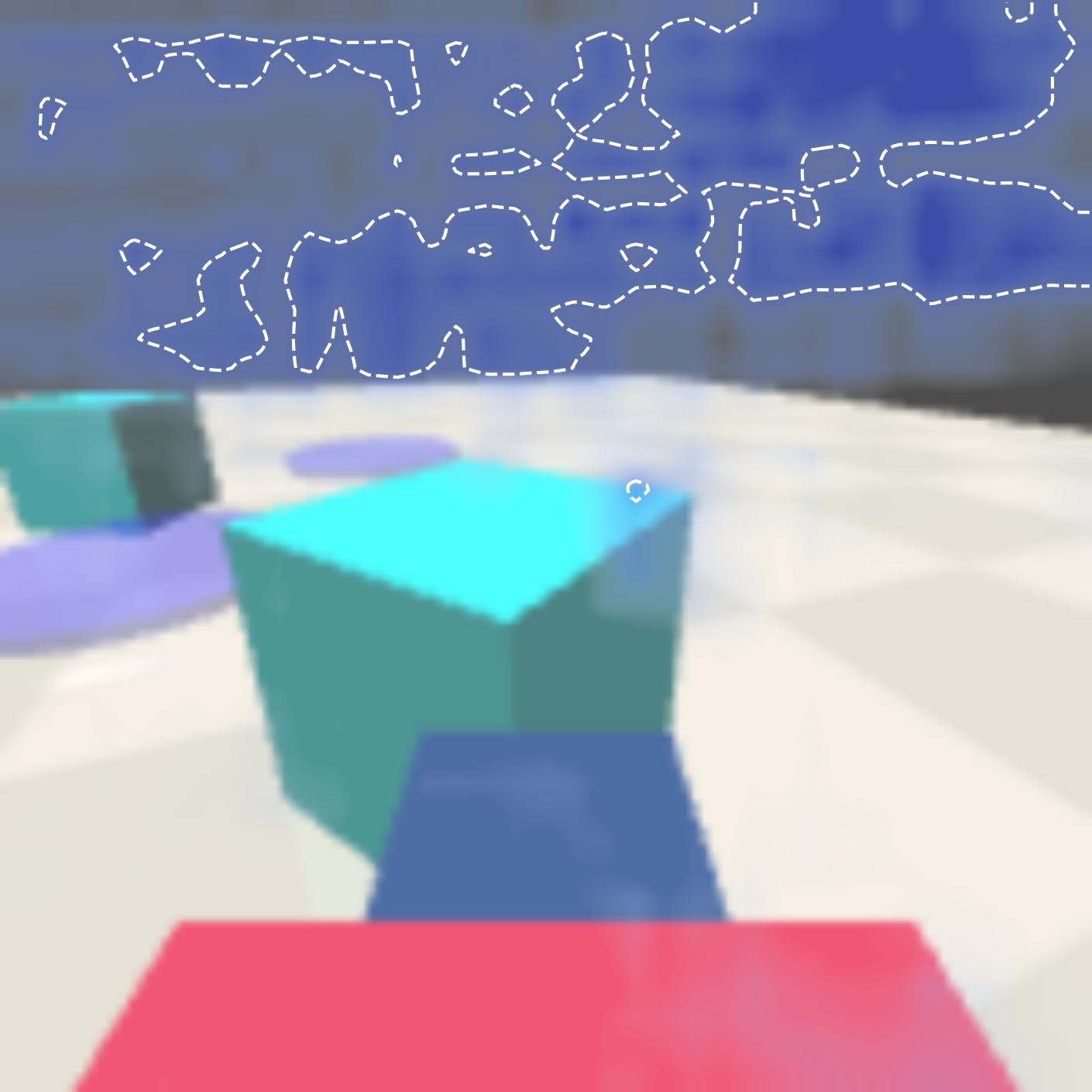}
    }
    \subcaptionbox{WM-R}{
    \includegraphics[width = .2\linewidth]{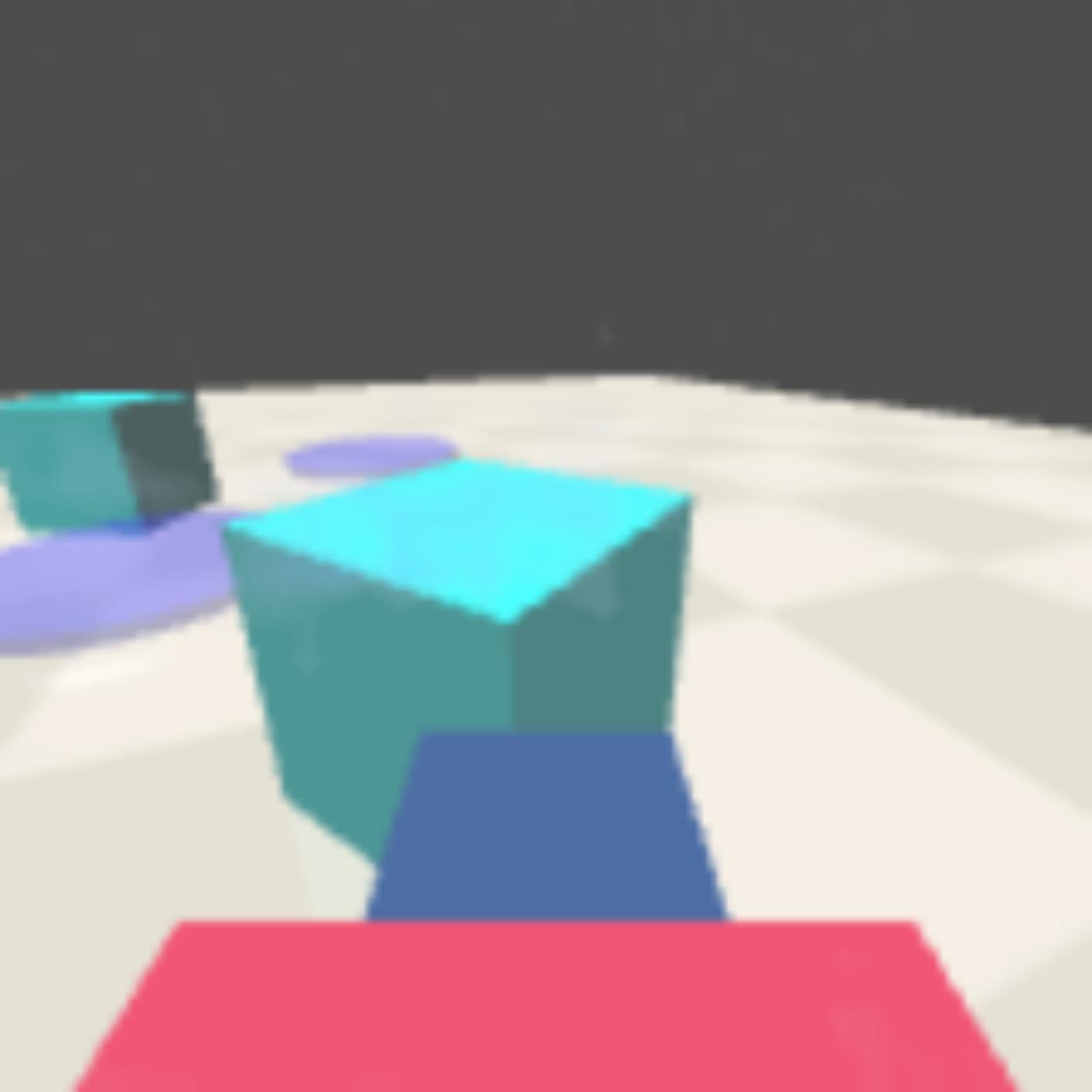}
    }
\caption{Saliency maps for CarGoal using by safety filters with different backbones; \textcolor{blue}{blue} overlay regions indicate areas where replacing pixels with white patches increases the HJ, making the latent state appear safer. Darker \textcolor{blue}{blue} signifies a higher increase in the HJ value upon replacing the patch. 
} 
\vspace{-0.4cm}
\label{fig:saliency}
\end{figure}

We observe that safety filters using DINO-CLS as the backbone, especially when the backbone is fine-tuned, consistently perform well across all tasks. Among filters with fine-tuned backbones, DINO-CLS-based filters achieve the lowest average number of violations on Dubins car and ManiSkill, and the third-lowest on CarGoal. Among filters with frozen backbones, they achieve the second-lowest number of violations on Dubins car and CarGoal, and the lowest on ManiSkill. As shown in  Figure~\ref{fig:saliency}, when obstacles appear in front of the car, only the DINO-CLS-based and VC1-based safety filters correctly focus on the front cube. Compared with the DINO-CLS-based filter, the VC1-based filter allocates more attention to the top side of the obstacle instead of the front side, which is more critical for the safety of the current state. As mentioned in Section~\ref{sec:vision_backbone}, DINO’s pretraining corpus includes depth-estimation datasets. We hypothesize that this exposure enables DINO to learn depth-aware representations, which are critical for assessing distance to obstacles.

Prior work reports VC1’s superiority on control benchmarks, outperforming R3M across general control tasks~\cite{vc1}. It is therefore surprising that in our evaluations, VC1 does not consistently achieve top-tier performance and occasionally underperforms R3M. This finding suggests that a PVR well-suited for control policies is not necessarily well-suited for safety filtering.

Another interesting finding is that, despite sharing the same visual backbone, safety filters using DINO as a backbone perform worse than those with DINO-CLS. We hypothesize that DINO’s representation—formed by concatenating all patch embeddings—introduces redundancy, inflates the dimensionality of the state space, and dilutes safety-relevant signals, making it harder to learn an HJ value function, which typically does not scale well beyond a few dimensions. In contrast, DINO-CLS relies on the CLS-token embedding, yielding a compact, semantically aggregated representation that is easier for the filter to exploit. Because DINO’s higher-dimensional representation requires larger models and more data to be effective, holding model size and training data fixed makes DINO-CLS better as a backbone for safety filtering.

\subsection{Critic-only, or dynamics lookahead for switching between the safe and the nominal policies ?}

Fine-tuning the backbones renders the pre-trained dynamics models ineffective because it alters their encoders. Although the dynamics model can be retrained with the fine-tuned encoder, this incurs long additional training time. This issue raises another question: if the dynamics model is not retrained after encoder fine-tuning, does a safety filter trained without fine-tuning the backbone and evaluated with the \texttt{dynamics lookahead} method outperform a filter with a fine-tuned backbone evaluated with the \texttt{Critic-only} method?

In both CarGoal and ManiSkill, safety filters with non-fine-tuned backbones using the \texttt{dynamics lookahead} method generally exhibit substantially more violations compared to their fine-tuned counterparts evaluated with the \texttt{Critic-only} method (Table~\ref{tab:closeloop}). In contrast, on the Dubins car task, safety filters with the \texttt{dynamics lookahead} method slightly outperform those that fine-tune the backbones and use the \texttt{Critic-only} method. Overall, these results demonstrate that fine-tuning the PVRs of the safety filters and evaluating them with the \texttt{Critic-only} method is generally better than using \texttt{dynamics lookahead} and not fine-tuning the PVRs.

\begin{figure}[h]
\centering
\vspace{-0.4cm}
\includegraphics[width=0.5\textwidth]{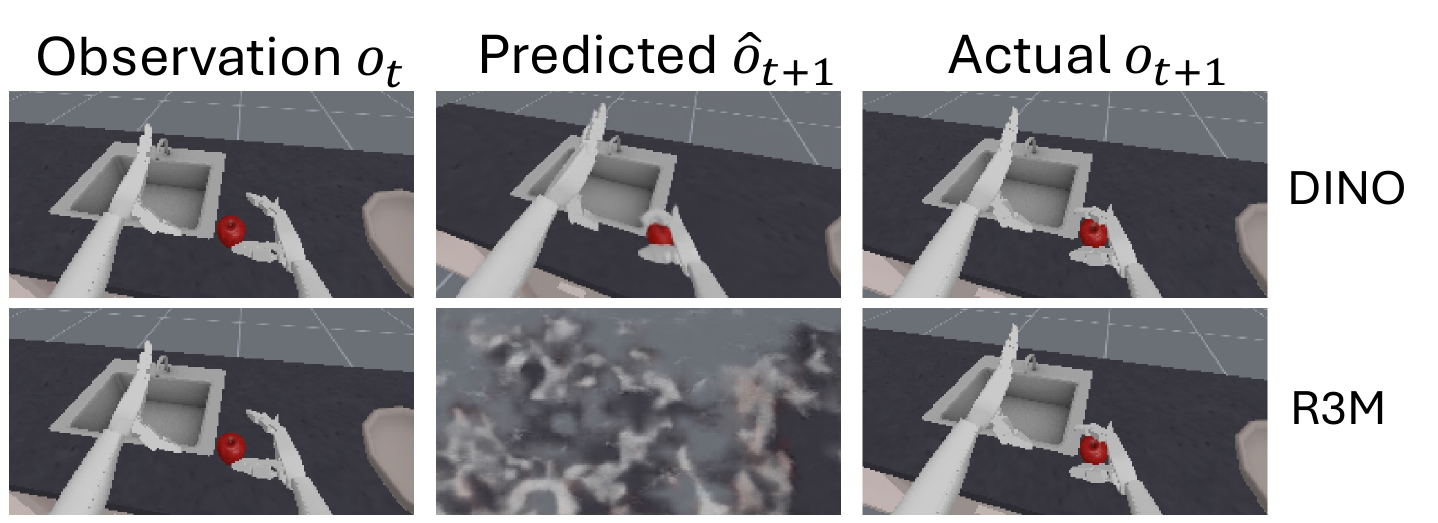}
\caption{Visual observations predicted using DINO-WM  }
\vspace{-0.3cm}
\label{fig:predicted_latent}
\end{figure}

The previous conclusion does not hold for all backbones. For example, the DINO backbone, whose representations concatenate patch embeddings leading to a very large vector representation of the size of 75284, the learned dynamics model is particularly accurate, as shown in Figure~\ref{fig:predicted_latent}. In these cases, its performance, as shown in Table \ref{tab:closeloop}, is consistently higher when evaluated with the dynamics model rather than the critic, indicating that when the backbone is well-suited to learning dynamics, the dynamics model might be the better choice.

On the other hand, when the backbone is not well-suited to learning dynamics, the critic is preferable. As shown in Figure~\ref{fig:predicted_latent}, the dynamics model’s predicted observations for R3M are worse than those for DINO, leading the safety filter with an R3M backbone to incur an average of 8.14 violations per episode, substantially higher than the 1.86 observed with DINO. When we use the \texttt{Critic-only} method instead of \texttt{dynamics lookahead}, the safety filter using R3M as a backbone achieves substantially fewer violations compared to that using DINO in both fine-tuned and non-fine-tuned PVR settings.

Given the drastic difference in R3M's performance using the \texttt{dynamics lookahead} and the \texttt{Critic-only} methods in ManiSkill (both fine-tuned and non-fine-tuned), we can deduce that R3M serves as an effective PVR for encoding safety-related features but is not well-suited as a representation model for predicting dynamics in the ManiSkill setup, in contrast to DINO. These observations highlight the importance of a PVR that not only supports effective feature extraction but also facilitates accurate next-state prediction.

\vspace{-0.05cm}
\subsection{Do we need a world model?}
\label{sec:do_we_need_worldmodel}
We observe in our closed-loop experiments that critic-based estimation of next-state HJ values for the fine-tuned PVR outperforms the dynamics-based approach on the non-fine-tuned PVR. If the dynamics is not used, and we can instead directly train an HJ while fine-tuning the backbone and using the critic to estimate the HJ value of the next state as in the \texttt{critic-only} method, why do we need to train a world model? 

For safety filters, simply encoding the visual observations and combining them (through concatenation or other techniques) with raw proprioception inputs to form the latent state is not sufficient. Recent work has demonstrated that fusing vision and proprioceptive information in a shared latent space outperforms simple concatenation of raw proprioceptive data with vision embeddings~\cite{Scaling_proprioceptive_visual_learning_withpvr}.  Wu et al.~\cite{wu2023daydreamer} achieve superior performance by encoding proprioceptive information through an MLP and aligning it with vision embeddings, compared to direct concatenation. 
Similarly, both Dreamerv3~\cite{dreamerv3} and DINO-WM~\cite{dinowm} incorporate dedicated proprioceptive encoders in their architectures. When training our world models, we also observed that naive concatenation of proprioceptive and action information with vision embeddings yields poor predictive performance for future proprioception in world model training. 

In our experiments, each environment provides proprioceptive inputs that provide the model with historical context (e.g., velocities in CarGoal, Carla, and ManiSkill), which cannot be inferred from a single observation. In settings without such proprioception information, the latent state must encode dynamics-related features such as velocities for the HJ value function to accurately estimate the BRS in latent space.
One can either concatenate representations from $H$ previous timesteps or use a Recurrent State-Space Model (RSSM) such as the Dreamerv3~\cite{dreamerv3} world model, which maintains a recurrent state that compresses historical observations. In this setting, the \texttt{Critic-only} method cannot be used without the world model, since the latent state depends on the recurrent state $r_t$, i.e., $z_t \sim \text{enc}_\theta(z_t \mid o_t, r_t)$ where $r_t$ is updated at every time step by the world model. 



\subsection{If the learned failure classifier is good, does it imply learning a good safety filter?}
Our results indicate that obtaining a strong failure classifier does not necessarily translate into an effective safety filter when using the same backbone. As an example, as seen in  Table~\ref{tab:corr}, we observe that DINO performs well on ManiSkill, Dubins car, and CarGoal, attaining one of the highest correlation scores when fine-tuned. However, when used as the backbone for training safety filters, which is shown in Table~\ref{tab:closeloop}, it does not achieve the lowest number of violations on any of these tasks, regardless of fine-tuning.

\subsection{Are PVR-based safety filters lightweight enough for deployment?}
In practice, inference speed and model size are critical, as they determine whether a model can be deployed in real-world applications and on edge devices. Table~\ref{tab:time} reports the total inference time and model size of the different filters. The maximum inference time of all the safety filters is $1.93\times10^{-2}$~s, which is sufficient for real-time operation at $50$~Hz. The largest model size of all the safety filters is $611$~MB, which fits within the storage capacity of most edge devices. Combining the above information, we conclude that the safety filters with PVR backbones can be deployed in real-world applications.
\vspace{-0.05cm}
\begin{table}[ht]
\centering
\scriptsize 
\setlength{\tabcolsep}{3pt} 
\renewcommand{\arraystretch}{1.2}

\resizebox{0.9\columnwidth}{!}{%
\begin{tabular}{lcccccc}
\toprule
& {DINO-CLS} & {DINO} & 
 {VC1} & {R3M} & 
  {Resnet} & {WM-R} \\
\midrule
\textbf{Inference Time (ms)} & 16.1 & 19.3 & 16.7 & 16.3 & 8.9 & 11.9 \\
\textbf{Model Size (MB)} & 291 & 510 & 611 & 275 & 275 & 275 \\
\bottomrule
\end{tabular}}
\caption{Inference time and model size for safety filters with different backbones. 
Model size includes the world model (encoders and predictor) and the safety filter. 
Inference time accounts for all steps: encoding the observation, predicting the next latent state, and computing safe action.}
\label{tab:time}
\vspace{-0.4cm}
\end{table}

\section{Conclusion}
We presented an evaluation of pre-trained visual representations (PVR) as perception backbones for safety filters. We compared several PVRs, considering frozen, fine-tuned, and scratch-trained variants across multiple simulated environments and analyzed their efficacy for both failure classification and safety filter learning. We explained the trade-offs between the PVR variants, studied whether one of the PVRs is superior, evaluated whether learned world models or Q-functions are better for switching decisions to safe policies, and discussed computational demands in training and deploying these models.
\vspace{-0.2cm}
\bibliographystyle{IEEEtran}
\bibliography{references}
\end{document}